\documentclass[]{spie}  

\usepackage{amsmath,amsfonts,amssymb}
\usepackage{graphicx}
\usepackage[colorlinks=true, allcolors=blue]{hyperref}
\usepackage{makecell}

\title{Coupling deep and handcrafted features to assess smile genuineness}

\author[a]{Benedykt Pawlus}
\author[a]{Bogdan Smolka}
\author[a]{Jolanta Kawulok}
\author[a]{Michal Kawulok}

\affil[a]{Faculty of Automatic Control, Electronics and Computer Science, Gliwice, Poland}

\begin{document} 
\maketitle

\begin{abstract}
Assessing smile genuineness from video sequences is a vital topic concerned with recognizing facial expression and linking them with the underlying emotional states. There have been a number of techniques proposed underpinned with handcrafted features, as well as those that rely on deep learning to elaborate the useful features. As both of these approaches have certain benefits and limitations, in this work we propose to combine the features learned by a long short-term memory network with the features handcrafted to capture the dynamics of facial action units. The results of our experiments indicate that the proposed solution is more effective than the baseline techniques and it allows for assessing the smile genuineness from video sequences in real-time. 

\end{abstract}


\section{INTRODUCTION}
\label{sec:intro}  


Identifying facial expressions in digital images and videos plays a crucial role in human face recognition~\cite{Kawulok2016FDA,WangDeng2021}, with numerous effective solutions proposed over the years. Most existing methods rely on the facial action coding system (FACS)~\cite{Ekman1978}, which describes facial activity as a combination of fundamental muscle movements known as action units (AUs). Notably, AUs can be accurately detected in facial images using computer vision techniques\cite{Baltruvsaitis2015}. Recent advancements in facial expression recognition focus not only on developing deep learning-based approaches~\cite{LiDeng2020} but also on enhancing our understanding of the link between facial expressions and human emotions~\cite{Duran2021}. A key challenge in this area is that expressions are often posed rather than occurring spontaneously~\cite{Straulino2023}. The ambiguity in the relationship between facial expressions and underlying emotional states may be intentional, as in cases of deception~\cite{Prome2024}, but can also arise naturally. A smile exemplifies this complexity—it can be deliberately posed, while a spontaneous smile may reflect various positive emotions such as pleasure, happiness, or joy, yet it can also emerge from sadness, fear, or embarrassment~\cite{LaFrance2011}.

There have been a number of attempts reported in the literature aimed at recognizing the genuineness of various facial expressions~\cite{JiaWang2021,ChengKotsia2018}, with the smile being given considerable attention. While it has been attempted to assess the smile genuineness from static images~\cite{Radlak2018}, the dynamics of displaying this facial expression are characterized with definitely higher discrimination capabilities~\cite{Zloteanu2018}. Therefore, the state-of-the-art approaches are mostly focused on analyzing video sequences to determine whether the smile is spontaneous or posed~\cite{Dibeklioglu2015,Wu2017,Kawulok2016Iberamia,Kawulok2021PLOS}. They include techniques underpinned with handcrafted features based on: (\textit{i})~the locations of facial landmarks~\cite{Dibeklioglu2015,Faroque2022}, (\textit{ii})~spatial-temporal textural features~\cite{Wu2017}, (\textit{iii})~dynamics of smile intensity~\cite{Kawulok2016Iberamia}, and (\textit{iv})~facial AUs~\cite{Kawulok2021PLOS}, as well as those that rely on deep learning to elaborate the useful features~\cite{YangHossain2020,Faroque2022}. While deep features extracted with convolutional neural networks (CNNs) commonly allow for achieving better classification scores, they are rather difficult to interpret~\cite{Arrieta2020} and computationally expensive. As both handcrafted and deeply-learned features have certain benefits and limitations, there have been many attempts to combine these two kinds of features to further improve the classification capabilities. Such approaches were elaborated for a variety of computer vision tasks~\cite{Duan2023,Nalepa2022ACIIDS}, including facial expression recognition~\cite{FanTjahjadi2019}. However, such approaches have not been studied so far for assessing smile genuineness and to our best knowledge this paper reports the first attempt to fill this gap.

Our contribution consists in proposing a solution that combines our earlier AU dynamics analysis (AUDA) with RealSmileNet that employs a CNN to extract deep features from subsequent frames of a video sequence, whose dynamics are analyzed with a long short-term memory (LSTM) network. At first, we process the AUDA features (which we made publicly available~\cite{Kawulok2020AUDA}) using an LSTM and we compare the achieved performance with the one obtained relying on the deep features. Furthermore, we study different fusion techniques to combine the handcrafted AUDA features with the deep features, which allows us to improve the overall classification score.

\section{PROPOSED APPROACH}

The proposed solution exploits the AUDA features, outlined in Section~\ref{sec:auda}, which are combined with the deep features extracted with the RealSmileNet encoding modules~\cite{YangHossain2020}, and classified using the LSTM architecture of RealSmileNet. This architecture, as well as the investigated approaches to combine these features, are explained in Section~\ref{sec:RealSmileNet}. 

\subsection{Handcrafted AUDA Features}\label{sec:auda}

First of all, we assess the AU intensities in each frame of a video sequence relying on the histogram-of-oriented-gradients (HOG) features classified using a support vector machine (SVM). Implementation of the SVM-HOG technique~\cite{Baltruvsaitis2015} that extracts 17 different AUs is available in the OpenFace library~\cite{Baltruvsaitis2016}. Also, we estimate smile intensity relying on an SVM classifier trained over local binary pattern features~\cite{Kawulok2016Iberamia}. For each AU feature, we retrieve the signal dynamics by applying linear regression in a sliding window of $\omega$ subsequent frames. This renders the trend line slope:
\begin{equation} \label{eq:slope}
\delta = \frac{\sum_{i=1}^\omega \left(t_i-\overline{t}\right)\left(v_i-\overline{v}\right)}{\sum_{i=1}^\omega \left(t_i-\overline{t}\right)^2} 
\end{equation}
and the regression coefficient:
\begin{equation} \label{eq:coeff}
r = \frac{\sum_{i=1}^\omega \left(t_i-\overline{t}\right)\left(v_i-\overline{v}\right)}{\sqrt{\sum_{i=1}^\omega \left(t_i-\overline{t}\right)^2 \sum_{i=1}^\omega \left(v_i-\overline{v}\right)^2}},
\end{equation}
where $v$ is the signal value, $t$ is the frame capture time, while $\overline{v}$ and $\overline{t}$ are the average values of $v$ and $t$ inside the window. When computing the AUDA features, two values of $\omega=\{9,27\}$ are used at video frequency of 50 frames per second (fps). Also, we smooth the $\delta$ signal by preparing its $r$-adjusted variant: $\hat{\delta}_i = \delta_i \left|r_i\right|$. Based on the dynamics, the smile phases are detected, namely: onset (when the smile appears), apex (when the smile intensity remains high), and offset (when the smile disappears). Furthermore, we capture the second-order dynamics by analyzing the $\delta$ signal as in Eq.~(\ref{eq:slope}).

The AUDA features are composed of three groups~\cite{Kawulok2021PLOS}: 154 frame-wise features, 119 AU-wise features that capture the dynamics of each AU (extracted from each of the aforementioned four phases), and 1088 cross-AU features that capture the mutual relations between the AU signals (also extracted from each of four phases). The cross-AU features are extracted from a signal $\delta_\Delta$ computed as the absolute difference between the dynamics of two AU signals. In addition to that, we compute the maximum and minimum values in the $r$-adjusted dynamic signals to retrieve the time difference between maximal increase and decrease of the two considered AU signals. The features extracted in this way can be subsequently classified to obtain the final decision on whether the smile is spontaneous or posed. In our earlier work~\cite{Kawulok2021PLOS}, we proposed an SVM ensemble that classifies them in a multi-level manner.

\subsection{Classification Schemes} \label{sec:RealSmileNet}

\begin{figure}
    \centering
    \includegraphics[height=0.33\textwidth]{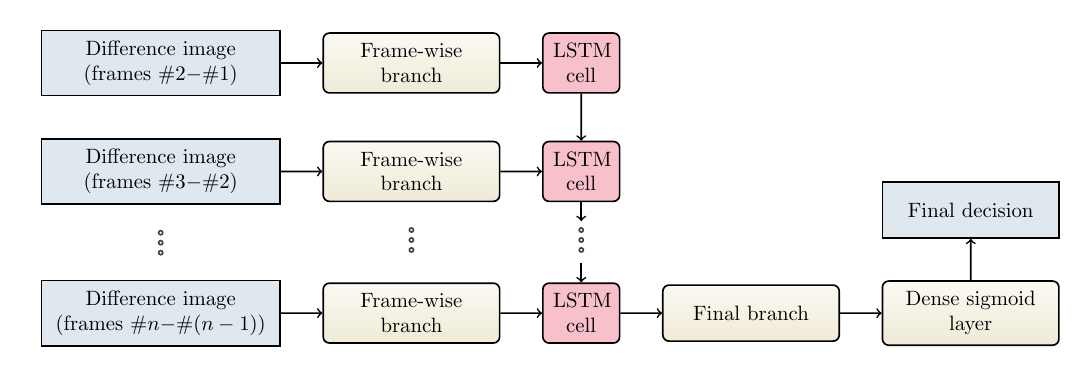}
    \caption{Outline of the RealSmileNet architecture. Each input image, being a difference between two consecutive frames, is processed with a frame-wise branch to extract deep features that are fed to a corresponding LSTM cell (red color). The output features of the last LSTM cell enter the classification block composed of the final branch and a dense sigmoid layer that retrieves the final decision on smile genuineness.}
    \label{fig:RealSmileNet}
\end{figure}

The RealSmileNet network~\cite{YangHossain2020} is composed of frame-wise branches, each of which extracts features from an image being a difference between two subsequent video frames, an LSTM module, and the final classification block (Figure~\ref{fig:RealSmileNet}). The extracted frame-wise feature vectors are fed to subsequent cells of the LSTM network, whose final outcome is processed with the final branch and classified with a dense sigmoid layer in the classification block. 

\begin{figure}
    \centering
    \includegraphics[height=0.33\textwidth]{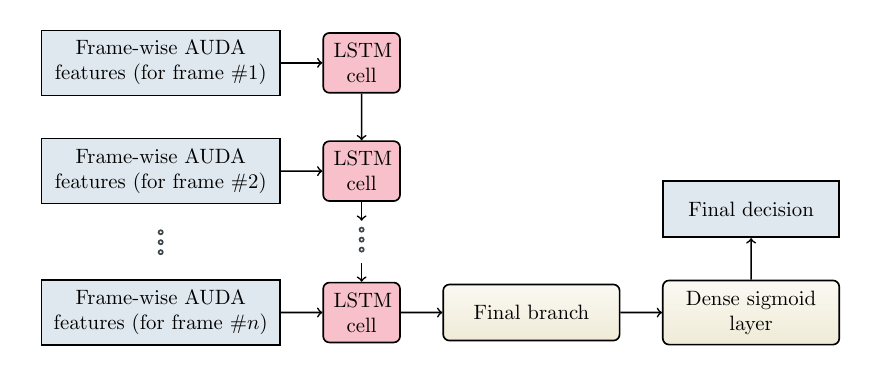}
    \caption{Outline of the frame-wise AUDA features classification scheme. The features extracted from each video frame are fed to a corresponding LSTM cell (red color). The output features of the last LSTM cell enter the classification block composed of the final branch and a dense sigmoid layer.}
    \label{fig:frame_wise}
\end{figure}

At first, we have adapted the RealSmileNet architecture to process both the frame-wise AUDA features, as well as the phase-wise features (AU-wise and cross-AU ones). The frame-wise features are fed directly to the LSTM cells in an analogous manner as the deep features extracted from each frame-difference image (Figure~\ref{fig:frame_wise}), and the network is trained to classify the smile based on the frame-wise AUDA features. 

\begin{figure}
    \centering
    \includegraphics[height=0.2\textwidth]{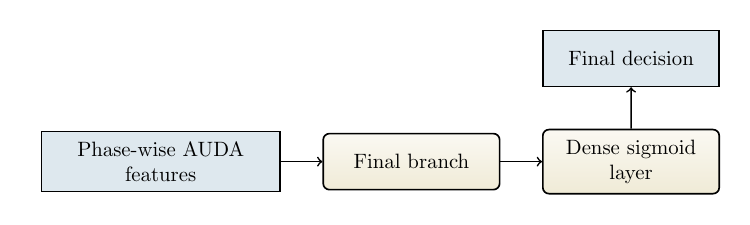}
    \caption{Outline of the architecture that classifies the phase-wise AUDA features (both AU-wise and cross-AU ones). The feature vector is fed to the classification block composed of the final branch and a dense sigmoid layer.}
    \label{fig:phase_wise}
\end{figure}

In contrast to the frame-wise features, the dimensionality of the AU-wise and cross-AU features does not depend on the number of frames in the sequence that shows a smile, so they are directly fed to the final classification block (Figure~\ref{fig:phase_wise}). For each of four phases, we have 119 AU-wise and 1088 cross-AU features, hence 476 and 4352 features for the whole sequence, respectively. However, as the number of AU-wise features is much smaller than of the cross-AU features, we train separate models for these two cases.

In order to benefit from both the handcrafted AUDA features and the deep features, we adopted a late-fusion strategy---we concatenate all the features extracted by the final branches of four models: (\textit{i})~the original RealSmileNet model (based on the deep features) and the models that process (\textit{ii})~the frame-wise AUDA features, (\textit{iii})~the AU-wise features, and (\textit{iv})~the cross-AU features. The feature vectors are concatenated and fed to the final dense sigmoid layer (same as in the classification block). Each model that contributes to the final fusion is trained beforehand, and the final dense sigmoid layer is trained to realize the fusion and classify the concatenated feature vector.

\section{EXPERIMENTS}

\begin{figure}[!t]
    \centering
    \includegraphics[width=\textwidth]{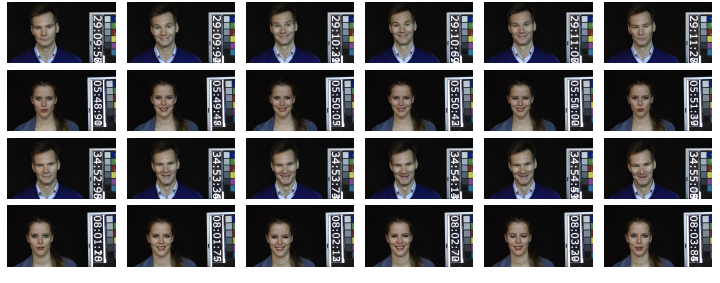}
    
    \caption{Examples of selected video frames from the UvA-NEMO dataset (for subjects no. 1 and no. 400) showing posed smiles (two upper rows) and spontaneous smiles (two lower rows).}
    \label{fig:uva-nemo}
\end{figure}

We have validated our approach over the UvA-NEMO benchmark dataset~\cite{Dibeklioglu2012} composed of 1240 video sequences of posed and spontaneous smiles (643 and 597 sequences, respectively, involving 400 subjects), captured at 50\,fps (see Figure~\ref{fig:uva-nemo}). The dataset\footnote{The UvA-NEMO database is available at \url{https://www.uva-nemo.org/index.html}.} is split into 10 folds, each of which contains recordings of different subjects, thus prepared for the 10-fold cross validation. We have exploited the AUDA features published in Ref.~\citenum{Kawulok2020AUDA}, and we used the official RealSmileNet~\cite{YangHossain2020} implementation published by the authors\footnote{Available at \url{https://github.com/Yan98/Deep-learning-for-genuine-and-posed-smile-classification}.}. The network was retrained $10\times$ to test each fold using the scripts provided by the authors. 

In Table~\ref{tab:scores}, we present the obtained quantitative results for RealSmileNet, for the AUDA features, as well as for the proposed feature fusion. The scores for RealSmileNet are lower than those reported in the original paper~\cite{YangHossain2020} (82.1\%) despite using the official code published by the authors---we have observed a strong overfitting to the training set which could be potentially reduced, however this was beyond the scope of the reported research. It can be seen that the AUDA features lead to higher classification scores, but their fusion improves the classification quality even more. In order to verify whether the deep features contribute to the increase in the classification quality, we have fused exclusively the AUDA features without the deep features. The result is slightly worse in that case, confirming that the use of deep features improves the score. In the table, we also report the times consumed to process the whole video sequence using RTX 2080 with 8\,GB VRAM---overall, it is clear that the investigated techniques allow for real-time processing. 
\begin{table}[b!]
    \caption{Classification accuracy (in \%) and processing times for the UvA-NEMO database obtained relying on the features extracted using individual techniques and based on the proposed feature fusion.}
    \centering
    \resizebox{\textwidth}{!}{
    \begin{tabular}{lcccccccccccc}

 \Xhline{2\arrayrulewidth}
    & \textbf{Time} & \multicolumn{10}{c}{\textbf{Fold}} & \\ \cline{3-12}
    \textbf{Method} & \textbf{[ms]} & \textbf{1} & \textbf{2} & \textbf{3} & \textbf{4} & \textbf{5} & \textbf{6} & \textbf{7} & \textbf{8} & \textbf{9} & \textbf{10} & \textbf{Average} \\
  \Xhline{2\arrayrulewidth}
RealSmileNet & $65.1$ &	$73.8$ & 	$72.0$ & 	$75.6$ & 	$72.8$ & 	$76.8$ & 	$75.0$ & 	$75.2$ & 	$72.8$ & 	$72.8$ & 	$77.4$ & 	$74.4\pm1.9$ \\
AUDA (frame-wise) & $2.3$ &	$77.8$ & 	$72.8$ & 	$82.7$ & 	$72.0$ & 	$84.0$ & 	$81.5$ & 	$75.2$ & 	$73.7$ & 	$75.2$ & 	$79.8$ & 	$77.5\pm4.3$ \\
AUDA (AU-wise) & $11.3$ &	$84.9$ & 	$80.0$ & 	$81.9$ & 	$83.2$ & 	$78.4$ & 	$82.3$ & 	$77.6$ & 	$76.3$ & 	$76.8$ & 	$79.8$ & 	$80.1\pm2.9$ \\
AUDA (cross-AU) & $81.8$ & $81.8$ & 	$73.6$ & 	$83.5$ & 	$84.0$ & 	$76.0$ & 	$82.3$ & 	$76.8$ & 	$73.7$ & 	$80.0$ & 	$82.3$ & 	$79.4\pm4.0$ \\ 
Fusion of AUDA features & $98.2$ &	$81.8$ & 	$77.6$ & 	$82.7$ & 	$82.4$ & 	$80.8$ & 	$79.0$ & 	$79.2$ & 	$78.1$ & 	$80.0$ & 	$83.9$ & 	$80.5\pm2.1$ \\
Fusion of all features & $164.5$ &  	$81.8$ & 	$79.2$ & 	$85.8$ & 	$84.0$ & 	$83.2$ & 	$82.3$ & 	$76.8$ & 	$78.1$ & 	$84.0$ & 	$83.1$ & 	$81.8\pm2.9$ \\


    \Xhline{2\arrayrulewidth}
    \end{tabular}
}
    \label{tab:scores}
\end{table}

\section{Conclusions}

In this paper, we have presented our study on assessing smile genuineness relying on handcrafted and deep features classified using a neural network. First of all, we showed that the handcrafted AUDA features which capture the dynamics of facial AUs, better discriminate between spontaneous and posed smiles than relying on the deep features extracted using convolutional layers from images being the differences between subsequent video frames. An important advantage of the AUDA features lies in their interpretability, as the classification decisions can be easily traced back to the individual AUs describing specific facial muscles activity. However, it is worth noting that exploiting the deep features during the fusion improves the scores, which indicates the potential of complementing the AU-based features with those learned from the data. Furthermore, our study may be helpful in designing more effective architectures of deep neural networks that would be more focused on analyzing the dynamics of specific facial regions that contribute the most to the discriminative capabilities of our AUDA features. Also, we plan to study the influence of image spatial and temporal resolution on the classification accuracy, including the possibility of applying super-resolution reconstruction~\cite{Tarasiewicz2021ICIP} before extracting the facial features. Such approaches were already applied to recognizing facial expressions~\cite{VoLee2020}, and they may also be valuable in assessing their genuineness.

\acknowledgments 
This work was funded by the National Science Centre, Poland, under Research Grant 2022/47/B/ST6/03009. BS and JK were supported by the Silesian University of Technology funds for developing and maintaining research potential.


\end{document}